# VinaFood21: A Novel Dataset for Evaluating Vietnamese Food Recognition

Thuan Trong Nguyen*, Thuan Q. Nguyen*, Dung Vo*, Vi Nguyen*, Ngoc Ho*,
Nguyen D. Vo[†], Kiet Van Nguyen[†], Khang Nguyen [†]
*University of Information Technology, Vietnam National University*, Ho Chi Minh, Viet Nam
Email: *{ 18521471, 18521470, 18520641, 18521636, 16520825}@gm.uit.edu.vn
[†]{nguyenvd, kietnv, khangnttm}@uit.edu.vn

*Abstract*—Vietnam is such an attractive tourist destination with its stunning and pristine landscapes and its top-rated unique food and drink. Among thousands of Vietnamese dishes, foreigners and native people are interested in easy-to-eat tastes and easy-to-do recipes, along with reasonable prices, mouthwatering flavors, and popularity. Due to the diversity and almost all the dishes have significant similarities and the lack of quality Vietnamese food datasets, it is hard to implement an auto system to classify Vietnamese food to make people easier to discover Vietnamese food. This paper introduces a new Vietnamese food dataset called VinaFood21, a dataset that consists of 13,950 images correspond to 21 dishes. We use 10,044 images for model training and 6,682 test images to classify each food in the VinaFood21 dataset and achieved an average accuracy of 74.81% in the testing set when fine-tuning CNN EfficientNet-B0.

## I. INTRODUCTION

Vietnamese food image recognition has become a promising computer vision application. Food recognition gives a possibility to automatically identify food items from each image captured by a camera, which makes people easier when discovering Vietnamese food. It is challenging and difficult because there are many similarities and diversities between different types of foods in three regions of the country, making it easy to confuse predictions even if humans cannot recognize them exactly. Specifically, the dishes taken can differ a lot in terms of ratio, lighting, background. Besides, the Vietnamese food dataset is still lacking in quantity, and almost all of them are not diverse dishes enough, so at the heart of this paper, we want to mention a larger Vietnamese food dataset we have created called VinaFood21. This dataset contains 13,950 images of 21 popular Vietnamese dishes. Besides, we present the methods we use to evaluate datasets. The highest results achieved an average accuracy of 74.81% when fine-tuning CNN EfficientNet-B0.

The contributions of our paper can be summarized as follows:

- We introduce a new Vietnamese diverse food image datasets.
- We experiment with multiple methods to conduct an extensive evaluation of our dataset.

The rest of this paper can be organized as follows. In section II: Summarization of related works. Section III: Describes the collecting, cleaning process, and detailed information of the dataset. Section IV: We propose methods for our problem. Section V: We analyze and evaluate the results achieved. The paper ends with a conclusion and gives some directions for future work.

## II. RELATED WORK

### A. Food Datasets

In recent years, food recognition has received more and more attention, so food datasets have been increasing in number rapidly. Table 1 summarizes statistics of publicly available datasets for food recognition. ETHZ Food-101 [1], with 101 most famous food from Western cuisines and approximately 1,000 images per class, includes very diverse and visually and semantically similar food classes such as Apple pie, Escargots, and Sashimi Onion rings, Macarons, etc. Vireo Food-172 dataset, containing food images from 172 Chinese food categories, covered eight primary foods, including Vegetables, Soup, Bean products, Egg, Meat, Seafood, Fish, and Staple. ChineseFoodNet, a larger Chinese cuisine dataset, contained 185,628 images from 208 food categories that the author collected from a top favorite Chinese food survey. NutriNet dataset includes 225,953 images from 520 food and drink classes but is limited to Central European food items. Overall, these datasets only belong to one country or regional cuisine. The FoodX-251 dataset, containing 158,000 images for 251 classes, which are fine-grained and visually similar, for example, different types of cakes, sandwiches, puddings, soups, and pasta. A large-scale and highly diverse food image dataset with 500 categories and about 400,000 images called ISIA Food-500, which covered various countries and regions, including Eastern and Western cuisines and existing typical ones mainly belong to the following 11 categories: Meat, Cereals, Vegetables, Fish, Fruits, Dairy, Bakery, Fats, Confectionary, Beverages, and Eggs.

TABLE I
*Summary Food datasets are available.*

| Dataset/Authors | Images | Categories | Coverage | Year |
|---|---|---|---|---|
| ETHZ Food-101 [1] | 101,000 | 101 | Western | 2014 |
| Vireo Food-172 [2] | 110,241 | 172 | Chinese | 2016 |
| ChineseFoodNet [3] | 185,628 | 208 | Chinese | 2017 |
| NutriNet [4] | 225,953 | 520 | Central European | 2017 |
| FoodX-251 [5] | 158,846 | 251 | Misc. | 2019 |
| ISIA Food-500 [6] | 399,726 | 500 | Misc. | 2020 |
| Thai *et al.* [7] | 2,315 | 5 | Vietnamese | 2017 |
| Ung *et al.* [8] | 8,903 | 13 | Vietnamese | 2020 |
| UEC Food100 [9] | 14,361 | 100 | Japanese | 2012 |
| UEC Food256 [10] | 25,088 | 256 | Japanese | 2014 |
| Sushi-50 [11] | 3,963 | 50 | Japanese | 2019 |
| KenyanFood13 [12] | 8,174 | 13 | Kenyan | 2019 |

Thai *et al.* built the Vietnamese cuisine dataset [7], including 2,315 images with only five signature Vietnamese dishes: Vietnamese roll cake, sizzling cake, broken rice, fried chicken, and Pho. Based on the existing Vietnamese food dataset, Ung *et al.* developed a new dataset [8] that contained 13 dishes within 8,903 images by adding some popular food in a daily day such as Vietnamese thick rice noodles, Vietnamese bread, Vietnamese Kuy teav, etc.

## B. Food Recognition

The food recognition problem is based on the number of labels to classify and each label's set of features. This feature set distinguishes it from the rest. This problem can describe as the input is an image of Vietnamese food, and the output is the name of that dish. (Figure 1)

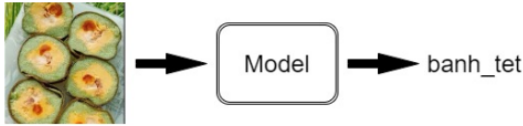

Fig. 1. Food recognition problem.

Zahisham *et al.* [13] used deep convolutional neural network (DCNN) architecture by fine-tuning a pre-trained model called ResNet50 and achieved accuracies of $41.08\%, 39.75\%$, and $35.32\%$ for the ETHZ Food-101 [1], UEC Food100 [9], and UEC Food256 [10] datasets, respectively. Another creative deep learning approach uses three CNN models together. Yanai and Kawano [14] proposed a method that combines fine-tuned AlexNet, GoogleNet and ResNet as an ensemble network called FoodNet with final achieved Top-1 accuracy 72.12% on ETHZ Food-101 [1] dataset. Thai *et al.* [7] fine-tune GoogleNet model with the best accuracy is 97.39%. Ung *et al.* [8] conducted experiments and achieved the highest performance which it gets the top $K$ accuracy ($K = 1, 3, 5$) as $92.0\%, 99.0\%$, and $99.78\%$ with InceptionResnetv2.

## C. Support Vector Machine (SVM)

SVM [15] is a linear classification algorithm that is one of the most commonly used algorithms in Machine Learning. That is, after training, we obtain the hyperplane, which will clearly divide the classes. The goal is to find a hyperplane so that the distance between the dividing line and the data points between classes is maximum (known as the maximum-margin hyperplane).

## D. Deep Learning Model

*1) VGG-19:* VGG [16] is a simple model with a deeper depth and was developed from AlexNet by adding convolutional layers to improve accuracy. VGG19 is a variant of the VGG model, which in short consists of 19 layers (16 convolution layers, 3 fully connected layers, 5 MaxPool layers, and 1 SoftMax layer). There are other variants of VGG like VGG11, VGG16, and others.

*2) ResNet-50:* ResNet [17] architecture consists of 2 characteristic blocks, Conv Block, and Identity Block. The model's accuracy will be saturated to a certain threshold or even make it less accurate. ResNet solved this by using a skip connection. The skip connection will connect from the previous layer to the next layer and skip some intermediate layers to reduce information loss. ResNet also is the earliest architecture that applies Batch Normalization. There are many variants of ResNet architecture with the same concept but with a different number of layers: ResNet-18, ResNet-34, ResNet-50, ResNet-110, ResNet-164, ResNet-1202, etc.

*3) DenseNet-121:* DenseNet [18] are quite similar to ResNets. However, DenseNet can solve the problem of vanishing gradients better than ResNet. DenseNet has a simple connectivity pattern to ensure the maximum flow of information between layers both in forward computation and backward gradient computation. This network connects all layers, so each layer obtains additional inputs from all preceding layers and passes its feature-maps to all subsequent layers. DenseNet-121 with only 8 million parameters, but it has higher precision than ResNet-50 with nearly 26 million parameters on the ImageNet dataset. DenseNet also has many variations such as DenseNet-121, DenseNet-169, DenseNet-201, DenseNet-264. Figure 2 illustrates the DenseNet network architecture.

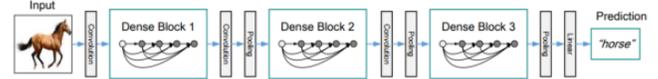

Fig. 2. DenseNet network architecture. [1]

*4) Inception-V3:* : Inception [19] has many popular versions are as follows: Inception-V1, Inception-V2 and Inception-V3, Inception-V4 and Inception-ResNet. Inception-V3 is a network architecture inheritance of Inception-V1. A batch normalization layer and a ReLu activation are added after the convolution layers. Inception-V3 solved representational bottlenecks. For example, the output's size dropped dramatically compared to the factorization method's input and efficient computation. Figure 3 illustrates the Inception network architecture.

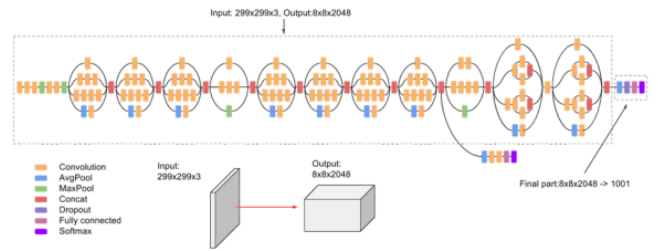

Fig. 3. Inception network architecture. [2]

*5) Xception:* Xception [20] proposed by Google is the model improved from Inception-V3, replaces standard modules with convolutional masses that can be separated in depth. Depthwise convolution is to perform convolution 3x3 on each input channel and combine the results. Then pointwise convolution, operate convolution 1x1 on the convolution results in depthwise convolution. The two operations' order is inconsistent: Inception performs a convolution 1x1, then a convolution of 3x3. Meanwhile, depthwise convolution does the opposite (this does not have a significant impact). In short, Xception improves model efficiency without increasing the network's complexity. Figure 4 illustrates the Xception network architecture.

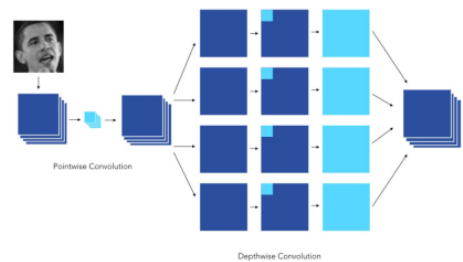

Fig. 4. Xception network architecture. [3]

*6) EfficientNet-B0:* EfficentNet [21] was introduced by Tan and Le, who studied the model scaling and identified that carefully balancing the depth, width, and resolution of the network can lead to better performance. They proposed a new scaling method that uniformly scales all dimensions of depth, width, and network resolution. They used the neural architecture search to design a new baseline network and scaled it up to obtain a family of deep learning models. There are 8 variations of EfficientNet (B0 to B7), and the simplest one is EfficientNet-B0, which has an outstanding result with Top-1 77% accuracy with $5,3$ million parameters. Figure 5 illustrates the EfficentNet network architecture.

---
[1] https://medium.com/intuitionmachine/notes-on-the-implementation-densenet-in-tensorflow-beeda9dd1504
[2] https://paperswithcode.com/method/inception-v3
[3] https://maelfabien.github.io/deeplearning/xception

Fig. 5. *EfficentNet network architecture.* [4]

### E. Feature Extraction

*1) Histograms of Oriented Gradients (HOG):* HOG [22] was proposed by Dalal and Triggs and announced at the CVPR 2005 Conference. HOG decomposes an image into small squared cells, computes a histogram of oriented gradients in each cell, normalizes the result using a block-wise pattern, and returns a descriptor for each cell.

*2) Deep learning models:* When performing deep learning feature extraction, we use the pre-trained models in II-D as an arbitrary feature extractor, allowing input images and stop at the specified layer and take the output of that layer as a feature.

## III. VINAFOOD21 DATASET

### A. Category Selection

Vietnamese cuisine encompasses diverse dishes from the mainstream culinary traditions in all three regions of Vietnam to the street food with original and creative recipes. We conduct a small survey on the Internet to choose the most favorite street food and dishes in typical traditional southern Vietnamese meals. Eventually, we statistics 21 dishes to continue to data collection for our dataset.

### B. Data Collection and Pre-processing

Most of the images were collected from social networking sites like Instagram, Facebook, Bing, Flickr posted by users with related tags. We expanded search terms by adding keywords, such as food and dish.

After this step, we collected more than $30,000$ color images, with no size limit. These images are saved in 21 folders for each dish. However, these images may be kept in an incorrectly labeled folder. Furthermore, images are low quality, distorted, and duplicate. So we do this by removing them and doing the labeling. The label set is assigned numbers $0 - 20$.

### C. Label Validation

In this phase, we perform two different sub-stages to check mistaken in data labeling. The two sub-stages are described as follows.
- Self-checking: Re-check your labels.
- Cross-checking: We cross-check each other's labels. If we find any errors in the dataset, we will discuss each other.

### D. Dataset Description

After completing the above works, VinaFood21 contains $13,950$ images are corresponding to 21 dishes. The total dataset size is $2GB$. The number of images per category is in the range of $(300, 1200)$. Figure 6 shows some samples per class.

VinaFood21 is a more comprehensive food dataset that surpasses existing Vietnamese Food datasets [7] [8] from the following three aspects. **(1) Larger data volume**: VinaFood21 has $13,950$ images from 21 dishes, which has created a new milestone for complex Vietnamese food recognition. **(2) More extensive category coverage**: It consists of 21 categories, which is about $2 \sim 3$ times that of existing Vietnamese food datasets. **(3) Higher diversity**: Food categories from our dataset cover various high intra-class distances (different look but similar type) and low inter-class distances (similar look but different type), including both street food and a daily meal. Moreover, our dataset is more challenging in various layouts, side dishes, and dishes in different lighting conditions.

## IV. METHODS

In this section, we evaluate the dataset on a variety of methods and present the metrics used. The whole process is shown in Figure 7.

### A. Implementation Process

Currently, the dataset is not large enough and increase the dataset's challenge, and we add 20% more binary images for each class. After this step, the number of obtained images is $16,732$. Next, the dataset was divided into training and testing sets with 60% and 40%, respectively. To build a training model, we approach the problem in two directions:

*1) SVM-based feature extraction:* In this work, rather than training a CNN from scratch, we used pre-trained CNN to feature extractions from a wide range of images. The network will remove the final layers (layers responsible for classifying), and only the final fully-connected (FC) layer is used to obtain the feature vector. Besides, HOG - a traditional feature extraction method, was also tested. To perform multiclass classification, SVM was utilized with features computed.

*2) Transfer Learning:* We implement transfer learning from different models trained on the ImageNet dataset to take advantage of the features learned by those models using deeper architectures and with more training time, specifically VGG19, ResNet50, DenseNet121, Inception-V3, Xception, and EfficientNet-B0. Transfer learning was implemented by loading the weights trained in the ImageNet dataset into each model and freezing each model's base layers while removing the top layers trained on the ImageNet classes, then replaced with trainable layers meant to learn classification on the VinaFooood21 dataset.

### B. Data Normalization

All dish images will be resized according to the applied models' input standards before the feature extraction calculation and the deep learning networks' input sizes. Specifically, in this article, the HOG feature and features extracted from deep learning models are VGG-19, ResNet-50, DenseNet-121, Inception-V3, Xception, EfficientNet-B0. Input image sizes 64x128 respectively for HOG feature extraction, 224x224 for deep learning networks: VGG-19, ResNet-50, DenseNet-121, EfficientNet-B0 and 229x229 are used for networks: Inception-V3, Xception. In this paper, we conduct experiments on $10,044$ images for training and $6,682$ images for testing. (Figure 8)

TABLE II
*Shapes of feature vectors.*

| Feature | Input shape | Output shape |
|---|---|---|
| HOG | (64, 128) | (1, 3780) |
| VGG-19 | (224, 224) | (1, 4096) |
| ResNet-50 | (224, 224) | (1, 2048) |
| DenseNet-121 | (224, 224) | (1, 1024) |
| Inception-V3 | (229, 229) | (1, 2048) |
| Xception | (229, 229) | (1, 2048) |
| EfficientNet-B0 | (224, 224) | (1, 1280) |

[4]https://ai.googleblog.com/2019/05/efficientnet-improving-accuracy-and.html

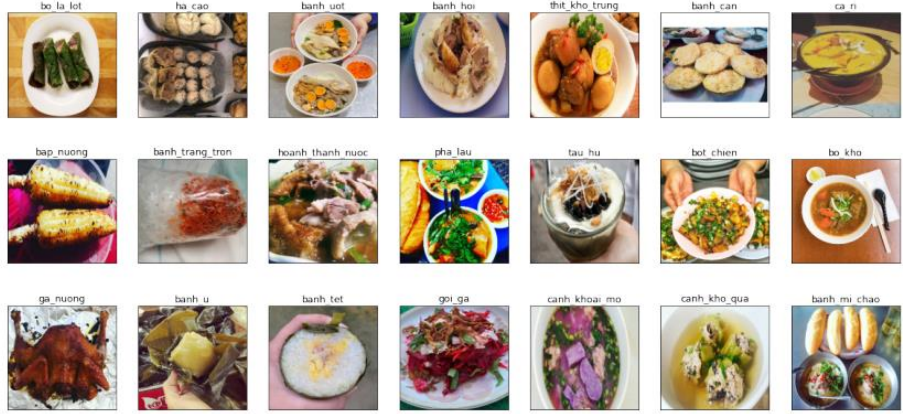

Fig. 6. Illustration 21 dishes in VinaFood21.

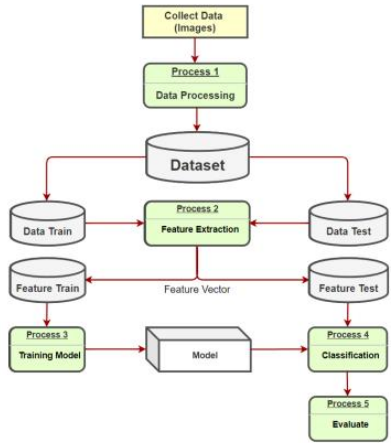

Fig. 7. The experimental process.

## C. Evaluation Metrics

*1) Accuracy:* indicates the fraction of predictions got right.

$$Accuracy = \frac{\sum_{i=1}^{C} \frac{TP_i + TN_i}{TP_i + TN_i + FP_i + FN_i}}{C}$$

*2) Precision:* indicates true positives probability in all positive predictions.

$$Precision = \frac{\sum_{i=1}^{C} \frac{TP_i}{TP_i + FP_i}}{C}$$

*3) Recall:* indicates the probability of truly being negative when the prediction is negative (contrast to precision).

$$Recall = \frac{\sum_{i=1}^{C} \frac{TP_i}{TP_i + FN_i}}{C}$$

*4) F1-score:* indicates the balance between precision and recall.

$$F1 - score = 2\frac{Precision \times Recall}{Precision + Recall}$$

Denoting $TP$ is True Positive, $FP$ is False Positive, $TN$ is True Negative, $FN$ is False Negative, and $C$ the number of classes.

## V. RESULTS AND DISCUSSION

### A. Experimental Results

After adjusting the dataset to each used method, we experimented according to the procedure in Figure 7, and the result is averaged over 21 classes. Experimental results on the testing set with the SVM method and classification results from the deep learning model are listed in Tables III, IV, respectively.

TABLE III
Experimental results with SVM classifier. The best performance is highlighted.

| Feature | Precision | Recall | Accuracy | F1-score |
|---|---|---|---|---|
| EfficientNetB0 | 76.03% | 74.34% | 73.99% | 74.91% |
| Xception | 73.07% | 72.22% | 71.54% | 72.40% |
| InceptionV3 | 72.31% | 70.78% | 70.46% | 71.30% |
| ResNet50 | 72.38% | 70.98% | 70.44% | 71.42% |
| VGG19 | 63.10% | 61.40% | 61.00% | 61.93% |
| DenseNet121 | 37.19% | 36.91% | 36.43% | 36.68% |
| HOG | 24.60% | 25.21% | 24.75% | 24.58% |

TABLE IV
Experimental results with Deep learning model. The best performance is highlighted.

| Model | Epochs | Model detail | Accuray |
|---|---|---|---|
| EfficientNet-B0 | 100 | fine-tuning training top N layers | 74.81% |
| EfficientNet-B0 | 100 | training last FC layer | 71.86% |
| Xception | 31 | training from scratch | 60.70% |
| Inception-V3 | 20 | training from scratch | 59.64% |
| VGG-19 | 10 | training from scratch | 42.85% |
| DenseNet-121 | 21 | training from scratch | 27.67% |
| ResNet-50 | 15 | training from scratch | 12.66% |

Table III shows the experimental results with the SVM model. We also experimented with multiple kernels and achieved the highest accuracy using the feature extracted from EfficientNet-B0 73.99% in the case that the SVM kernel is linear. Specifically *bap-nuong* is the dish with the highest prediction result (precision: 95.35%, recall: 86.01%, f1 score: 90.44%) and the most incorrectly predicted dish is *ca-ri* (precision: 47.62%, recall: 46.78%, f1-score: 47.20%). Linear SVM has the lowest results when using the HOG feature 24.75%, even if the *banh-uot* has the highest results but is not considered good (precision: 33.11%, recall: 49.26%, f1-score: 39.60%) and the lowest result is *canh-khoai-mo* (precision: 16.39%, recall: 13.99%, f1-score: 15.09%). Based on SVM results, we find that CNN EfficientNet-B0 can give better results on our problem, so we decided to conduct more experiments with EfficientNet-B0. First, all the layers are frozen except the last FC layer, we get average accuracy is 71.86% with *bap-nuong* (precision: 91.85%, recall: 86.71%, f1-score: 89.21%) give the best prediction, and *ca-ri* (precision: 44.09%, recall: 32.75%, f1-score: 37.58%) has the worst prediction. Second, freezing the base layer and fine-tuning the top layer as shown in Figure 10b with dropout rates of 0.5 and 0.2. We achieved average accuracy of 74.81% with the highest prediction is *bap-nuong* (precision: 87.16%, recall: 90.21%, f1-score: 88.66%) and with the most incorrect prediction *ca-ri* (precision: 53.23%, recall: 38.6%, f1-score: 44.75%). Simultaneously, we try training from scratch CNN model in a few epochs, the best accuracy achieved with CNN Xception on the testing set 60.70% (Table IV). But, from then on, the model started to overfitting.

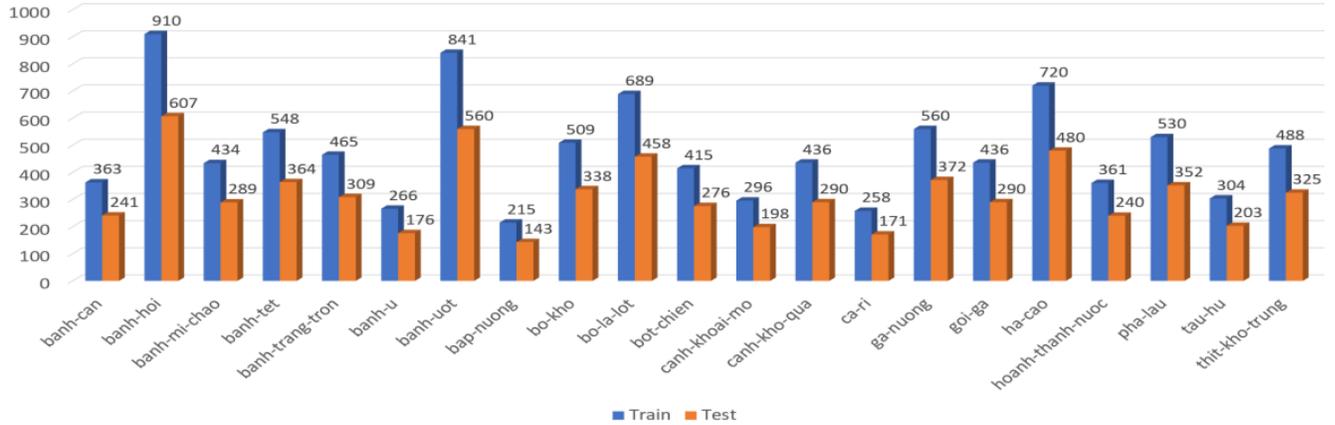

Fig. 8. *Organization of experimental data.*

the last two layers to achieve better results. Basically, CNN learns to make their domain objects linearly separable in the layer just before the last layer. Then, the last layer act as a linear classifier. When we adjust the final layer's knowledge to our dataset, this means just adjusting the linear classifier's weights. Because CNN is trained on the ImageNet dataset, so their domain is entirely different from the VinaFood21 dataset. Therefore, when we apply it to the VinaFood21 dataset, the dish may not be separated linearly in the last $2nd$ layer. However, when the adjusted CNN starts at the last $2nd$ layer, they will learn how to transform feature vectors created from the previous layer to be linearly separable in the last $2nd$ layer. From there, the dishes become linearly separable before being classified. On the other hand, transfer learning solves our problems more efficiently than training from scratch. In the ImageNet dataset, we found 10 food-related classes: apple, banana, broccoli, burger, egg, french fries, hot dog, pizza, rice, and strawberry. With transfer learning, our model takes advantage of previously learned knowledge instead of training from scratch when the VinaFood21 dataset is not large enough and encounters CNN models with the number of parameter numbers in the millions.

### C. Error Analysis

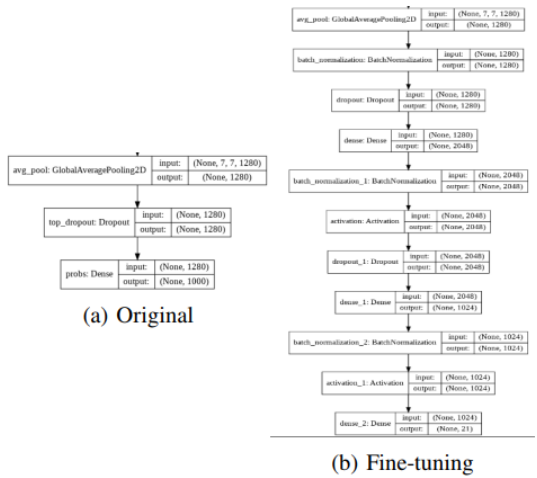

(a) Original

(b) Fine-tuning

Fig. 9. *Comparison between the top $N$ layer of EfficienNet-B0 original and fine-tuning.*

### B. Experimental Results Analysis

There were 3 notable results with CNN EfficientNet-B0 (Table III, IV). First, we use the feature extracted from CNN with SVM. Second, all layers are frozen except for the last FC layer that is retrained. Third, freeze all the base layers and add 2 FC layers before the final FC layer compared to the original EfficientNet-B0. The results show that CNN knowledge must be adjusted starting from

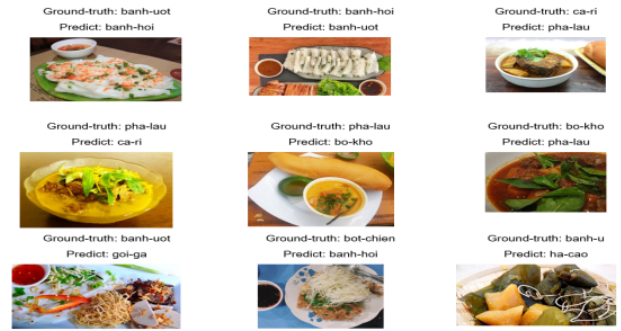

Fig. 10. *Incorrect predictions.*

By analyzing the model's predictive results on the testing set, we found that most dishes have similarities in color and layout very easily confused with each other (Figure 10). Besides, when looking at the noise data versus the total amount per dish, we found that the large percentage of noise data could cause the low model results (Figure 11). Therefore, these are considered the challenges of the VinaFood21 dataset.

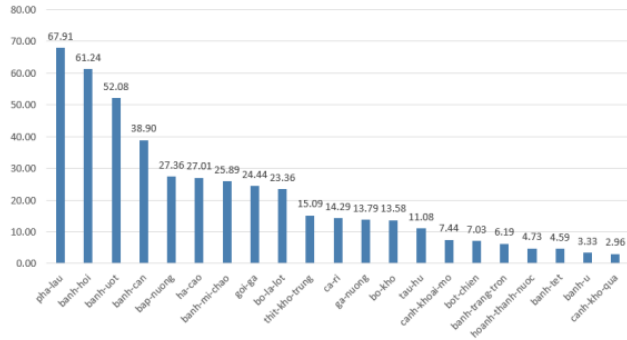

Fig. 11. *Statistical noise data (%)*

## VI. CONCLUSION AND FUTURE WORK

This paper introduced a new dataset for Vietnamese food with 21 types of popular dishes in Vietnam, VinaFood21. The dataset contains nearly 14,000 food images taken in everyday life. We have conducted extensive experiments using both the CNNs model and CNNs based on features to build the classification system. Based on the experimental results, fine-tuning a pre-trained model called EfficientNet-B0 outperforms the other comparison methods and achieved accuracies of 74.81%. In the future, we plan to collect more samples to extend the current dataset, expand the number of dishes, continue improving food recognition models, and try implementing an application on mobile devices. We hope to publish the VinaFood21 dataset and our methods to contribute to the research community in the Vietnamese food recognition problem.

## ACKNOWLEDGMENT

The research team would like to express our sincere thanks to the Multimedia Communications Laboratory (MMLab) and Faculty of Information Science and Engineering — University of Information Technology — Vietnam National University — Ho Chi Minh City, for supporting my team in this research process.

## REFERENCES

[1] L. Bossard, M. Guillaumin, and L. Van Gool, "Food-101 – mining discriminative components with random forests," in *Computer Vision – ECCV 2014*, D. Fleet, T. Pajdla, B. Schiele, and T. Tuytelaars, Eds., Cham: Springer International Publishing, 2014, pp. 446–461.

[2] J. Chen and C.-w. Ngo, "Deep-based ingredient recognition for cooking recipe retrieval," in *Proceedings of the 24th ACM International Conference on Multimedia*, ser. MM '16, Amsterdam, The Netherlands: Association for Computing Machinery, 2016, pp. 32–41. [Online]. Available: https://doi.org/10.1145/2964284.2964315.

[3] X. Chen, Y. Zhu, H. Zhou, L. Diao, and D. Wang, "Chinesefoodnet: A large-scale image dataset for chinese food recognition," May 2017.

[4] S. Mezgec and B. Koroušić Seljak, "Nutrinet: A deep learning food and drink image recognition system for dietary assessment," *Nutrients*, vol. 9, no. 7, 2017. [Online]. Available: https://www.mdpi.com/2072-6643/9/7/657.

[5] P. Kaur, K. Sikka, W. Wang, S. Belongie, and A. Divakaran, *Foodx-251: A dataset for fine-grained food classification*, Jul. 2019.

[6] W. Min, L. Liu, Z. Wang, Z. Luo, X. Wei, X. Wei, and S. Jiang, *Isia food-500: A dataset for large-scale food recognition via stacked global-local attention network*, 2020. arXiv: 2008.05655 [cs.CV].

[7] V.-P. Thai, T. Dang, Q. Pham, N. Pham, and B. Nguyen, "Vietnamese food recognition using convolutional neural networks," Oct. 2017, pp. 124–129.

[8] H. T. Ung, T. X. Dang, P. V. Thai, T. T. Nguyen, and B. T. Nguyen, "Vietnamese food recognition system using convolutional neural networks based features," in *Computational Collective Intelligence*, Cham: Springer International Publishing, 2020, pp. 479–490.

[9] Y. Matsuda and K. Yanai, "Multiple-food recognition considering co-occurrence employing manifold ranking," in *Proceedings of the 21st International Conference on Pattern Recognition (ICPR2012)*, 2012, pp. 2017–2020.

[10] Y. Kawano and K. Yanai, "Automatic expansion of a food image dataset leveraging existing categories with domain adaptation," in *Computer Vision - ECCV 2014 Workshops*, L. Agapito, M. M. Bronstein, and C. Rother, Eds., Cham: Springer International Publishing, 2015, pp. 3–17.

[11] J. Qiu, F. P. W. Lo, Y. Sun, S. Wang, and B. Lo, "Mining discriminative food regions for accurate food recognition," in *30th British Machine Vision Conference 2019, BMVC 2019, Cardiff, UK, September 9-12, 2019*, BMVA Press, 2019, p. 158. [Online]. Available: https://bmvc2019.org/wp-content/uploads/papers/0839-paper.pdf.

[12] M. Jalal, K. Wang, S. Jefferson, Y. Zheng, E. O. Nsoesie, and M. Betke, "Scraping social media photos posted in kenya and elsewhere to detect and analyze food types," in *Proceedings of the 5th International Workshop on Multimedia Assisted Dietary Management*, ser. MADiMa '19, Nice, France: Association for Computing Machinery, 2019, pp. 50–59. [Online]. Available: https://doi.org/10.1145/3347448.3357170.

[13] Z. Zahisham, C. P. Lee, and K. M. Lim, "Food recognition with resnet-50," in *2020 IEEE 2nd International Conference on Artificial Intelligence in Engineering and Technology (IICAIET)*, 2020, pp. 1–5.

[14] K. Yanai and Y. Kawano, "Food image recognition using deep convolutional network with pre-training and fine-tuning," in *2015 IEEE International Conference on Multimedia Expo Workshops (ICMEW)*, 2015, pp. 1–6.

[15] C. Cortes and V. VAPNIK, "Support-vector networks," *Chem. Biol. Drug Des.*, vol. 297, pp. 273–297, Jan. 2009.

[16] S. Liu and W. Deng, "Very deep convolutional neural network based image classification using small training sample size," pp. 730–734, 2015.

[17] K. He, X. Zhang, S. Ren, and J. Sun, "Deep residual learning for image recognition," in *2016 IEEE Conference on Computer Vision and Pattern Recognition (CVPR)*, 2016, pp. 770–778.

[18] G. Huang, Z. Liu, L. van der Maaten, and K. Weinberger, "Densely connected convolutional networks," Jul. 2017.

[19] C. Szegedy, V. Vanhoucke, S. Ioffe, J. Shlens, and Z. Wojna, "Rethinking the inception architecture for computer vision," Jun. 2016.

[20] F. Chollet, "Xception: Deep learning with depthwise separable convolutions," Jul. 2017, pp. 1800–1807.

[21] M. Tan and Q. Le, "EfficientNet: Rethinking model scaling for convolutional neural networks," in *Proceedings of the 36th International Conference on Machine Learning*, K. Chaudhuri and R. Salakhutdinov, Eds., ser. Proceedings of Machine Learning Research, vol. 97, Long Beach, California, USA: PMLR, Sep. 2019, pp. 6105–6114. [Online]. Available: http://proceedings.mlr.press/v97/tan19a.html.

[22] N. Dalal and B. Triggs, "Histograms of oriented gradients for human detection," in *2005 IEEE Computer Society Conference on Computer Vision and Pattern Recognition (CVPR'05)*, vol. 1, 2005, 886–893 vol. 1.